\documentclass[11pt]{article}
\usepackage{PRIMEarxiv}

\usepackage[utf8]{inputenc} % allow utf-8 input
\usepackage[T1]{fontenc}    % use 8-bit T1 fonts
\usepackage{hyperref}       % hyperlinks
\usepackage{url}            % simple URL typesetting
\usepackage{booktabs}       % professional-quality tables
\usepackage{amsfonts}       % blackboard math symbols
\usepackage{nicefrac}       % compact symbols for 1/2, etc.
\usepackage{microtype}      % microtypography
\usepackage{lipsum}
\usepackage{fancyhdr}       % header
\usepackage{graphicx}       % graphics
\graphicspath{{media/}}     % organize your images and other figures under media/ folder

\usepackage{amsmath}     % AMS Math package; equation environments, math addons

  \usepackage{amsthm}    % AMS theorem-like environments
  \usepackage{amssymb}   % AMS math symbols and styles like \mathbb/cal/frak/...
\usepackage{siunitx}     % Unified number/unit handling
\usepackage{booktabs}    % Better table control
\usepackage[capitalize]{cleveref}    % Smart referencing
  \crefname{section}{Sec.}{Secs.}
  \crefname{appendix}{App.}{Apps.}

\usepackage[font=small]{caption}    % Caption control
\usepackage[shortlabels,inline]{enumitem}  % Customize list envronments
\usepackage{multicol}  % Short multi-column environment
\usepackage{mathtools} % Fine-tune math typesetting
\usepackage{tensor}    % Indices for GR/QFT
\usepackage{cancel}    % Crossing out of expressions
\usepackage{subdepth}  % Alignment for sub/superscripts
\usepackage{adjustbox} % Resize large objects
\usepackage{tikz}      % Graphics magic

\usepackage{etoolbox}% http://ctan.org/pkg/etoolbox

\usepackage[numbers, sort&compress]{natbib}

\usepackage{blindtext}
\usepackage{verbatim}
\usepackage{color}
\usepackage{subfigure}

\definecolor{darkblue}{rgb}{0,0,0.4}

\hypersetup{
    colorlinks=true, %set true if you want colored links
    linktoc=all,     %set to all if you want both sections and subsections linked
    linkcolor = darkblue,  %choose some color if you want links to stand out
    citecolor = blue
}

\usepackage{pdfpages}

\AtEndEnvironment{theorem}{\null\hfill\qedsymbol}%

\newcommand{\beq}{\begin{equation}}
\newcommand{\eeq}{\end{equation}}
\newcommand{\ba}{\begin{array}}
\newcommand{\ea}{\end{array}}
\newcommand{\bea}{\begin{eqnarray}}
\newcommand{\eea}{\end{eqnarray} }
\newcommand{\be}{\begin{eqnarray}}
\newcommand{\ee}{\end{eqnarray} }
\newcommand{\bal}{\begin{align}}
\newcommand{\eal}{\end{align}}
\newcommand{\bi}{\begin{itemize}}
\newcommand{\ei}{\end{itemize}}
\newcommand{\ben}{\begin{enumerate}}
\newcommand{\een}{\end{enumerate}}
\newcommand{\bc}{\begin{center}}
\newcommand{\ec}{\end{center}}
\newcommand{\bt}{\begin{table}}
\newcommand{\et}{\end{table}}
\newcommand{\btb}{\begin{tabular}}
\newcommand{\etb}{\end{tabular}}

\newcommand{\nn}{\nonumber}

\DeclareMathOperator{\Tr}{Tr}

\providecommand{\ie}{\emph{i.e.}}
\providecommand{\eg}{\emph{e.g.}}

\providecommand{\RR}{\mathbb{R}}

\providecommand{\grad}{\textsl{g}}

\providecommand{\ie}{\emph{i.e.}}
\providecommand{\eg}{\emph{e.g.}}

\providecommand{\gvec}[1]{\boldsymbol #1}

\providecommand{\RR}{\mathbb{R}}

\providecommand{\mb}{\mathcal{B}}

\def\({\left(}
\def\){\right)}

\def\La{\mathcal{L}}

\def\mb{\mathcal{B}}

\def\bs{\boldsymbol}

\def\nn{\nonumber}

\usepackage{tabularx}
\usepackage{siunitx}
\usepackage{booktabs}
\usepackage{amsthm}
\newtheorem{thm}{Theorem}

\providecommand{\grad}{\textsl{g}}

\title{Noise Injection as a Probe of Deep Learning Dynamics}

\author{
Noam Levi\thanks{Both authors contributed equally to this work.} \\
Raymond and Beverly Sackler School of Physics and Astronomy\\
Tel-Aviv University\\
Tel-Aviv 69978, Israel \\
\texttt{noam@mail.tau.ac.il} \\
\AND
Itay M. Bloch$^*$ \\
Berkeley Center for Theoretical Physics\\
University of California, Berkeley, CA 94720\\
\texttt{itay.bloch.m@gmail.com}\\
\And
Marat Freytsis \\
NHETC, Department of Physics and Astronomy\\
Rutgers University\\
Piscataway, NJ 08854, USA\\
\texttt{marat.freytsis@rutgers.edu} \\
\AND
Tomer Volansky \\
Raymond and Beverly Sackler School of Physics and Astronomy\\
Tel-Aviv University\\
Tel-Aviv 69978, Israel \\
\texttt{tomerv@post.tau.ac.il} \\
}

\begin{document}
\maketitle

\begin{abstract}

  We propose a new method to probe the learning mechanism of Deep Neural Networks (DNN) by perturbing the system using Noise Injection Nodes (NINs).
  These nodes inject uncorrelated noise via additional optimizable weights to existing feed-forward network architectures, without changing the optimization algorithm.
  We find that the system displays distinct phases during training, dictated by the scale of injected noise.
  We first derive expressions for the dynamics of the network and utilize a simple linear model as a test case. 
  We find that in some cases, the evolution of the noise nodes is similar to that of the unperturbed loss, thus indicating the possibility of using NINs to learn more about the full system in the future. 
\end{abstract}

\section{Introduction} % (fold)
\label{sec:intro}

Deep learning has proven exceedingly successful, leading to dramatic improvements in multiple domains. 
Nevertheless, our current theoretical understanding of deep learning methods has remained unsatisfactory.
Specifically, the training of DNNs is a highly opaque procedure, with few metrics, beyond curvature evolution~\cite{hochreiter1997flat,sagun2016eigenvalues,guy1,ghorbani2019investigation, yao2020pyHessian,papyan2018full,li2020Hessian}, available to describe how a network evolves as it trains. 

An interesting attempt at parameterizing the interplay between training dynamics and generalization was explored in the seminal work of Ref.~\cite{DBLP:journals/corr/ZhangBHRV16}, which demonstrated that when input data was corrupted by adding random noise, the generalization error deteriorated in correlation with its strength. 
Noise injection has gained further traction in recent years, both as a means of effective regularization~~\cite{graves2011practical, ba2013adaptive, goodfellow2013challenges, srivastava2014dropout, wan2013regularization, kang2016shakeout, wager2013dropout, li2018whiteout, 10.1007/978-3-030-64221-1_16,https://doi.org/10.48550/arxiv.2102.07379}, as well as a route towards understanding DNN dynamics and generalization.
For instance, label noise has been shown to affect the implicit bias of Stochastic Gradient Descent (SGD)~\cite{blanc2020implicit,damian2021label,haochen2021shape,DBLP:journals/corr/abs-2110-06914,https://doi.org/10.48550/arxiv.2206.09841}, as sparse solutions appear to be preferred over those which reduce the Euclidean norm, in certain cases. 

In this work, we take another step along this direction, by allowing the network to actively regulate the effects of the injected noise during training.  
Concretely, we define \emph{Noise Injection Nodes} (NINs), whose output is a random variable, chosen sample-wise from a given distribution.
These NINs are connected to existing feed-forward DNNs via trainable \emph{Noise Injection Weights} (NIWs).
The network is subsequently trained to perform a classification/regression task using vanilla SGD. 

\begin{figure}
% [ht!]
\includegraphics[width=1\linewidth]{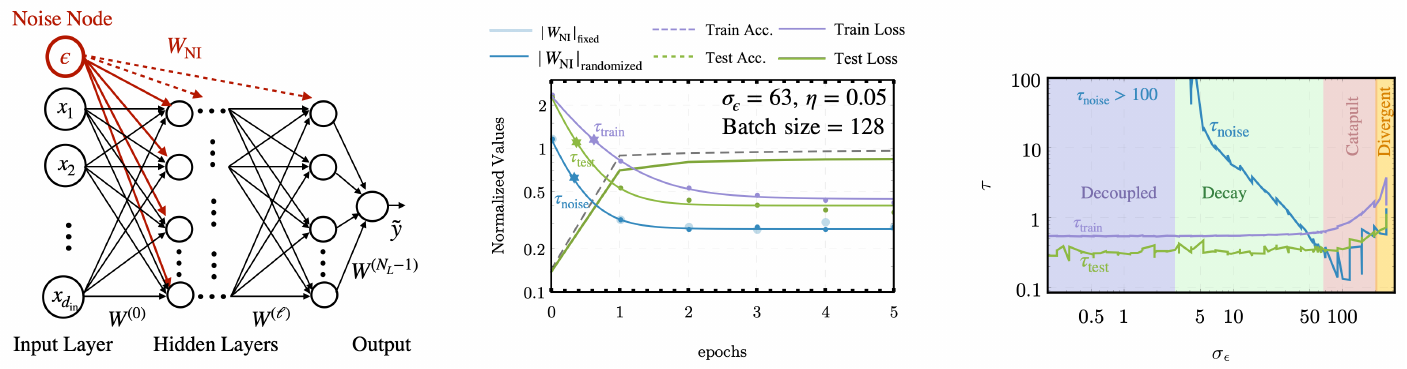}
\caption{
Schematic of a generic DNN, with the addition of a single NIN connected via NIWs.
 {\bf Center:}  
 Example evolution displaying a decay behavior for both the NIWs and the losses within the decay phase of the system discussed below and with a fixed noise strength, $\sigma_\epsilon$.  
 Two cases are shown: re-initialized NIN at every epoch, and fixed value NIN.  
 The similar behavior of the systems in both cases ({\bf blue} and {\bf light blue} points respectively) hint at a potent relation between the NIWs evolution and generalization. 
 The \textbf{blue}, \textbf{green} and \textbf{violet} stars indicate the NIW, test loss and training loss decay time-scales.
 The three solid curves are fits to exponential decays, while the data is represented with points.
\textbf{Right:}  The different decay times as a function of the noise injection magnitude.  The four shaded regions indicate the four phases of the system discussed in \cref{sec:linear_model}. 
The results are shown for a 3-hidden layer ReLU MLP with CE loss, trained on FMNIST to \SI{100}{\percent} training accuracy. 
}
\label{fig:probe}
\end{figure}

We study such systems both numerically and analytically, providing a detailed analytic understanding for a simple linear network.  
Our main results, partly summarized in \cref{fig:probe}, are as follows:

\begin{enumerate}[label=(\roman*)]
  \item The system exhibits 4 NIN-related phases, depending mostly on the strength of the injected noise.
  \item In two of the phases, the NIWs evolve to small values, implying that a well-trained network is able to recognize that the noise contains no useful information within it. \label{item:item_2}
  \item For those two phases, the NIW dynamics is dictated by the local curvature of the training loss function\footnote{
  Recently, the relationship between the inherent noise present in SGD optimization and generalization has been of interest, see~\cite{thomas2020interplay,roberts2021sgd} and references therein. Our work may be considered as a modeling of said SGD noise.
  }.
%%%%%%%%%%%%%%%%%%%%%%%%%%%%%%%%%%%%%%%%%%%%%%%%
\end{enumerate}

Item~\cref{item:item_2} may be expected if the NIN is re-randomized at each training epoch, yet we find essentially the same behavior repeated even when the NIN values are generated only once and fixed before training, putting them on equal footing with actual data inputs, as shown in~\cref{fig:probe} (center,right).
It appears that while the system might in principle be able to memorize the specific noise samples, optimization dynamics still prefer to suppress them.
This implies a relation between the NIN reduction mechanism and the network's ability to generalize, to be explored further in future works.

%%%%%%%%%%%%%%%%%%%%%%%%%%%%%%%%%%%%%%%%%%%%%%%%%%%%%%%%
%%%%%%%%%%     FIGURE 1 %%%%%%%%%%%%%%%%%%%%
%%%%%%%%%%%%%%%%%%%%%%%%%%%%%%%%%%%%%%%%%%%%%%%

\section{Noise Injection Weights Evolution}
\label{sec:theory}
Consider a DNN with parameters $\bs{\theta}= \{W^{(\ell)},b^{(\ell)}\in \RR^{d_{\ell} \times d_{\ell+1}},\RR^{d_{\ell+1}}| \ell=0, \dotsc ,N_L-1\}$, corresponding to the weights and biases, and $N_L$ layers, defined by its single sample loss function $\La :\RR^{d_\mathrm{in}} \to \RR$ and optimized under SGD to perform a supervised learning task.
Here, at each SGD iteration, a mini-batch $\mb$ consists of a set of labeled examples, $\{ (\bs x_i, \bs y_i) \}_{i=1}^{|\mb|} \subseteq \RR^{d_\mathrm{in}} \times \RR^{d_\mathrm{label}} $. 

We study the simple case of connecting a given NIN to a specific layer, denoted as $\ell_{\rm NI}$, via a NIW vector $W_\mathrm{NI} \in \RR^{1\times d_ {\ell_\mathrm{NI}+1} }$ (see \cref{fig:probe}, left).
In this setup, the injected noise is taken as a random scalar variable, $\epsilon$, sampled repeatedly at each SGD training epoch from a chosen distribution.
The NIWs' evolution is best studied via their effect on preactivations, defined as $\bs{z}^{(\ell)}=W^{(\ell)}\cdot \bs{x}^{(\ell)}+b^{(\ell)} $. When a NIN is added, the preactivations at layer $\ell_\mathrm{NI}$ are subsequently shifted to $\bs{z}^{(\ell_{\rm{NI}})}\to \bs{z}^{(\ell_\mathrm{NI})} + W_{\rm{NI}} \epsilon$.

For a single NIN connected at layer $\ell_\mathrm{NI}$, the batch-averaged loss function can be written as a series expansion in the noise translation parameter\footnote{In practice, it is often the case that one uses piece-wise analytic activation functions such as ReLU, and so if the noise causes the crossing of a non-analytic point, the above formal expansion is invalid. This subtlety does not change any of our conclusions and empirically we recognize the same phases when using ReLU activations.
}
\begin{align}
    L({\bs\theta}, {W_\mathrm{NI}})
    &=
    \frac{1}{|\mb|} 
    \sum_{\substack{ \{\bs x, \epsilon, \bs y\}
    \in \mb}}
        \La(\tilde{\bs \theta}; 
        \bs{z}^{(\ell_{\rm{NI}})}  +
        { W_{\rm{NI}}}\epsilon, \bs{y})
        \\ \nn
        &=
        L({\bs{\theta}})
        +
        \frac{1}{|\mb|} 
      \sum_{ \substack{ \{\bs x, \epsilon, \bs y\}
    \in \mb}}
          \sum_{k=1}^{\infty}
          \frac{(\epsilon { W^T_{\rm{NI}}} \cdot \nabla_{{\bs z^{(\ell_{\rm{NI}})}}}
              )^k }{k!}
        \La(\bs \theta; {\bs{x}}, \epsilon, {\bs y}),
\end{align}
%%%%%%%%%%%%%%%%%%%%%%%%%%%%%%%%%%%%%%%%%%%%%%%%%%%%%%%%%%%%%%%%%%%%%%
%%%%%% MIGHT PUT DERIVATION HERE INSTEAD OF APPENDIX FOR ARXIV %%%%%%%
%%%%%%%%%%%%%%%%%%%%%%%%%%%%%%%%%%%%%%%%%%%%%%%%%%%%%%%%%%%%%%%%%%%%%%
where ${\tilde{\bs \theta}}=\bs\theta \setminus\{W^{(\ell_{\rm{NI}})}\}$ and $L (\gvec{\theta})$ is the loss function in the absence of a NIN.
Focusing on a distribution with zero mean for the NIN (\eg, $\epsilon \sim \mathcal{N}(0,\sigma_\epsilon^2)$) and performing the batch averaging on each term, we arrive at the update rule for the NIWs from the noisy loss expansion\footnote{Additional $\sigma_\epsilon/\sqrt{|\mb|}$ corrections coming from the variances of the even terms emerge from batch-averaging, and are assumed to be negligible throughout this work.}
\begin{align}
\label{eq:newweight}
  W_{\rm{NI}}^{(t+1)} 
  = 
  W_{\rm{NI}}^{(t)}
              - \eta \frac{\sigma_\epsilon \Phi}
              {\sqrt{| \mb|}}
                  \sqrt{\langle (\bs\grad_{\ell_{\rm{NI}} }^{(t)})^2 \rangle}
              - \frac{\eta\sigma^2_\epsilon}{2}
                   \left\langle \mathcal{H}_{\ell_{\rm{NI}}}^{(t)} \right\rangle W_{\rm{NI}}^{(t)}
                   +\dots .
\end{align}
Here, batch averaging is denoted by $\langle \,\dotsi \rangle$, $\Phi$ is a random variable with zero mean and unit variance, and $\sigma^2_\epsilon$ is the variance of the injected noise. 
We denote the network-dependent {\it local} gradient and Hessian at the NIN layer as 
\begin{align}
{\bs{\grad} }_{\ell _{\rm{NI}}} 
 =  \nabla_{{\bs z}^{(\ell_{\rm{NI}})}} \La({\bs \theta}; {\bs x}, {\bs y}),
 \qquad
 { {\mathcal{H}_{\ell_{\rm{NI}} }} }  
=  { \nabla_{\bs z^{(\ell_{\rm{NI}})}}\nabla^T_{\bs z^{(\ell_{\rm{NI}})}}} \La(\bs \theta; \bs x, \bs y).
\end{align}
A more complete derivation, along with a proof for the $1/\sqrt{\mb}$ scaling of odd terms in the expansion is given in~\cref{apd:epsilon}.

Terminating the  expansion in \cref{eq:newweight} at 2nd order need not be valid for large $\sigma_\epsilon$.  
We thus proceed by studying a linear test case for which the 2nd order expansion is precise.
The persistence of analogous network behavior, and in particular its phases, for a more realistic setup is confirmed empirically.

\section{ Linear Toy Model}
\label{sec:linear_model}

Consider a two-layer DNN with linear activations and layer widths ($d_{0,1}=1$) and no biases ($b=0$), tasked with univariate linear regression\footnote{
We use this toy model as a proxy for a diagonal linear network~\cite{https://doi.org/10.48550/arxiv.1806.00468}. A more general treatment including width and depth effects will be presented in future work.}, and with a single NIN connected to the first layer ($\ell_\mathrm{NI}=0$). 
The data consists of a set of training samples $\{ ( x_i, y_i ) \in \RR\times \RR \}_{i=1}^m$, and we sample $x_i$ and the noise $\epsilon_i$ from the normal distributions, $x_i, \epsilon_i \sim \mathcal{N}(0,\sigma_{x,\epsilon}^2)$.
The corresponding data labels are given by a linear transformation of the inputs $y_i = M \cdot x_i$ with a fixed $M\in \RR$.  
This regression problem is solved by minimizing the empirical loss, taken as the Mean Squared Error (MSE), 
\begin{align}
  L_\mathrm{MSE}
    = \frac{1}{2|\mb|} \sum_{i\in\mb} ( w^{(1)} (w^{(0)} \cdot x_i + w_{\rm{NI}} \epsilon_i) - y_i )^2,
\end{align}
with optimal solution $w^{(1)}_* w^{(0)}_* = M, w_{\mathrm{NI},*}=0$.

The evolution of the system can be studied by focusing on the coupled SGD equations for the hidden layer weight and the NIW, parameterized as
\begin{align}
\label{eq:linear_evolution}
     w^{(1)}_{t+1} &= A_t \sigma_\epsilon
                      + w^{(1)}_t (1 - B_t\sigma_\epsilon^2) - C_t, 
                      \\
                      \nn
      w_{\mathrm{NI}, t+1} &= \tilde{A}_t \sigma_\epsilon
                              + w_{\mathrm{NI},t} (1-\tilde{B}_t \sigma_\epsilon^2) .
\end{align}

Here, the various terms are given explicitly by\footnote{
We match the local gradient $\sqrt{\langle {\bs \grad}_{0}^2\rangle} = \sigma_x w_t^{(1)} ( w_t^{(1)}w_t^{(0)} - M )$ and Hessian $\langle \mathcal{H}_0^{(t)}\rangle = 2 (w^{(1)}_t)^2 $ to \cref{eq:newweight}.
}
\begin{equation}
  \begin{split}
    A_t &= \eta \frac{\Phi_t \sigma_x}{\sqrt{|\mb|}} w_{\mathrm{NI},t}
    (2w_t^{(1)}w_t^{(0)} - M), \qquad
    \tilde{A}_t = \eta \frac{\Phi_t \sigma_x}{\sqrt{|\mb|}}
                   w_t^{(1)} (w_t^{(1)}w_t^{(0)} - M), 
                   \\  
    B_t & = \eta\,  w_{\rm{NI},t}^2, \qquad
    \tilde{B}_t = \eta(w^{(1)}_t)^2, \qquad
    C_t = \eta (w^{(1)}_t w^{(0)}_t - M) w^{(0)}_t \sigma_x^2,
  \end{split}
\end{equation}
and are functions of the NIW, the data weights, the learning rate $\eta$ and the batch size $|\mb|$.
 
 %%%%%%%%%%%%%%%%%%%%%%%%%%%%%%%%%%%%%%%%%%%%%%
 %%%%%%%% Figure 2 %%%%%%%%%%
\begin{figure}
  \centering
  \includegraphics[width=1\textwidth]{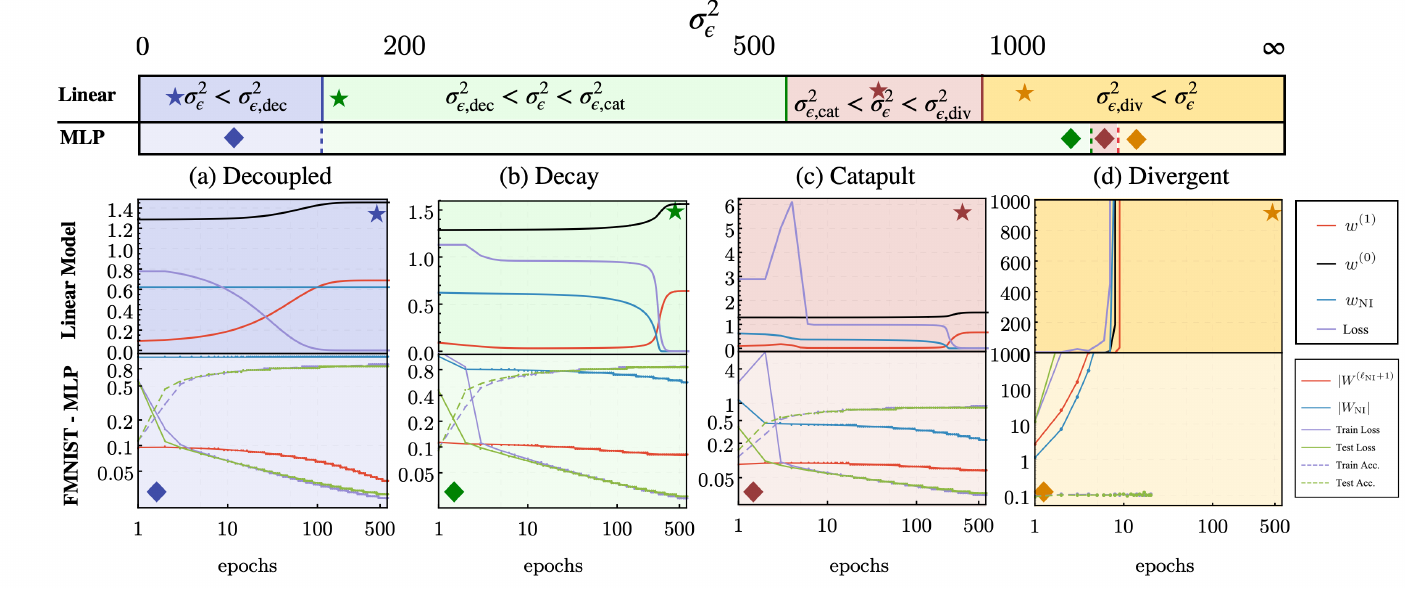}
  \caption{
  Evolution for a DNN with a NIN coupled to the first hidden layer, for two different models. 
  \textbf{Top bar:} Regions for the phases as a function of the noise strength $\sigma_\epsilon$.
  The \textbf{top darker shaded} and \textbf{bottom lighter shaded} bar regions correspond to the linear and Multi-Layer Perceptron (MLP) models respectively.
  The \textbf{stars} and \textbf{diamonds} indicate the values of $\sigma_\epsilon$ used in the bottom plots. 
  \textbf{Top plots:} Numerical solutions for the  various phases of \cref{eq:linear_evolution}, describing the linear model.
  The data weights (\textbf{black} and \textbf{red} lines), NIW (\textbf{blue}) and loss functions (\textbf{purple}) are shown.
  In the decoupled and decay phases, data weights approach optimum values while the NIW decays slowly,
  whereas a larger noise magnitude results in a longer relaxation time.
  In the initial stages of the catapult phase no learning is achieved until the NIW sufficiently decays, when standard learning is resumed.
  In the divergent phase the system fails to learn, with breakdown after only a few epochs.
  \textbf{Bottom plots}: Various phases of NIW dynamics during training for a 1-hidden layer MLP with MSE loss and ReLU activations, trained on FMNIST.
  \textbf{Solid} (\textbf{dashed}) curves represent the loss (accuracy) values, for training (\textbf{purple}) and test (\textbf{green}) instances.  
 The behavior displayed by the loss function, the NIWs (\textbf{blue}) and the subsequent layer weights (\textbf{red}) norms, corroborates the predictions of the linear model, discussed in \cref{sec:theory}.
 For experimental details, see \cref{apd:architecture_details}. 
  }
\label{fig:linear_phases}
\end{figure}

While non-linear, the solution to \cref{eq:linear_evolution} is rather simple and is dictated by $\sigma_\epsilon$ and the initial conditions.  
We identify four different phases, demonstrated in \cref{fig:probe} (right) and \cref{fig:linear_phases}: 

  {{\textit{Decoupled phase}}}.~
    When the scale of injected noise is sufficiently small, $\sigma_\epsilon\leq
    \sigma_{\epsilon, \rm{dec}}
    \equiv 2{\tilde{A}_0}/{\tilde{B}_0}$, the original optimization trajectory, driven by the $C_t$ term, is on average unaffected by the NIN, and the NIWs evolve according to a random walk, with step size dictated by the local gradient, $\tilde A_t$\footnote{
    While in principle the random walk could allow the system to exit this phase (if $w_{\rm NI,t}$
    sufficiently grows), this phase is quickly restored due to $\tilde B_t$. 
    }.

\textit{Decay phase}.~
For $\sigma_{\epsilon, \mathrm{dec}}^2 < \sigma_\epsilon^2 \ll \sigma_{\epsilon, \mathrm{cat}}^2  \equiv \min[2/B_0, 2/\tilde{B}_0]$ the dynamics is initially dominated by the $\tilde B_t$ and $C_t$ terms, and as a consequence $w_\mathrm{NI}$ exponentially falls.
Once $w_\mathrm{NI}$ is sufficiently small, the $A$-terms dominate and the dynamics is decoupled from the noise as in the previous phase.
  
The above two phases occur for rather small noise injection and, as discussed above, the noise is a mere small perturbation to the dynamics.     
However, for $\sigma_\epsilon>\sigma_{\epsilon, \rm cat}$, the equations become noise-dominated at initialization and 
the early  dynamics becomes insensitive to the original learning objective.
Early time evolution can then be understood by neglecting the $A$- and $C$-terms, and the equations become noise dominated at initialization, with the early times evolution entirely determined by $B_t,\tilde{B}_t$ as 
\begin{align}
    w^{(1)}_{t+1}
      =
      w^{(1)}_t
      (1- B_t\sigma_\epsilon^2  ) 
        , 
        \\ \nn 
        w_{\rm{NI },  t+1}
      =
      w_{\rm{NI},t} (1-\tilde{B}_t \sigma_\epsilon^2)
      .
\end{align}
%%%%%%%%%%%%%%%%%%%%%%%%%%%%%%%%%%%%%%%%%%%%%%%%%%%%%%%
%%%%%%%%%%%%%%%%%%%%%%%%%%%%%%%%%%%%%%%%%%%%%%%%%%%%%%%
Consequently, the equations describe a DNN, trained using completely random data with no labels or learning objectives. The network evolution can then be separated into the following two distinct phases:

\textit{Catapult phase}.~
When $\sigma_{\epsilon, \mathrm{cat}}< \sigma_\epsilon < \sigma_{\epsilon, \mathrm{div}} \equiv  \max(2/B_0, 2/\tilde{B}_0)$, some, but not all, of the network weights begin diverging, as a stiff equation regime ensues due to the discrete nature of the SGD algorithm.
Generally, this divergence can be driven by either the data or the NIWs, though in realistic scenarios the latter is more common and we therefore discuss for concreteness the $2/\tilde B_0<\sigma_\epsilon < 2/B_0$ case. 
While the NIW (and the loss function) diverges, $w_t^{(1)}$ decays fast enough for the Hessian to be reduced, allowing for the network to recover, and resulting in a catapult effect\footnote{We borrow the name from~\cite{lewkowycz2020large}, as the behavior of the system in this phase closely matches the large learning rate behavior of SGD, discussed in their work.}. 
%%%%%%%%%%%%%%%%%%%%%%%%%%%%%%%%%%%%%%%%%%%%%%%%%%%%%%%
%%%%%%%%%%%%%%%%%%%%%%%%%%%%%%%%%%%%%%%%%%%%%%%%%%%%%%%
At this point the dynamics of the system behave similarly to the Decay Phase, albeit typically at a slower rate since the Hessian is now significantly smaller.
These results are visible in \cref{fig:linear_phases}(c).

\textit{Divergent phase}.
Once $\sigma_\epsilon^2 \geq \sigma_{\epsilon,\rm{div}}^2$, the NIN overwhelms the network, and all weights (and loss function) diverge resulting in a failed training process.
In non-linear networks, this phase can also occur if the second order perturbative approximation for the loss function breaks down\footnote{
In theory, there may be intermediate steps in unique scenarios (\eg, the NIW decays, but without effect on the data weights' evolution), however we have not found these cases to empirically matter and ignore them here.
}.

% \nl{
While the discussion above pertained to a simplified scenario, it is expected to capture the features of the dynamics in the limit of large batch size $|\mb|\to \infty$ and small noise variance $\sigma_\epsilon \ll1$, as in this limit the second-order truncation of \cref{eq:newweight} is justified; we verify this numerically.
% }
The bottom row of \cref{fig:linear_phases} demonstrates the persistence of our predicted phase diagram for an over-parameterized DNN trained on the FMNIST dataset~\citep{xiao2017/online}\footnote{
All experiments were performed on a 20 node cluster, each consisting of 24-48 CPUs using Intel$^\circledR$ Xeon$^\circledR$ E5-2650 v4 CPU @ \SI{2.20}{\giga\hertz}, with further implementation details given in~\cref{apd:architecture_details}.
}.
While the decoupled and decay phases occur for similar $\sigma_\epsilon$ values as those in the linear model, the large noise behavior is altered by the choice of non-linearities, loss function, batch size and number of layers. 
In particular, the noise variance values dictating the phase boundaries are somewhat different than those predicted by the linear model, however the existence of the boundaries themselves persists throughout our experiments\footnote{We find that for bounded activations (\eg, sigmoid or tanh), the divergent phase does not occur, but rather neuron saturation causes the learning process to fail.}.
Some of these effects are partially explored in \cref{apd:more_exp}.

\section{Relating the Evolution of the NIWs and Loss}
As discussed in \cref{sec:intro}, during the decay and catapult phases, the NIWs decay as the network attempts to learn, a phenomena which repeats both for re-randomized as well as fixed value NINs.
Since the latter can be interpreted as the suppression of uncorrelated data features, understanding the relationship between the NIWs and loss evolutions could provide future insights regarding generalization.

Here, we take a step towards connecting the NIW dynamics with the loss dynamics, for the simple linear model, hinting that more complicated constructions could allow a more direct relationship between the two.
With no NINs and for small enough learning rates, 
\begin{align}
    \delta L^{(t+1)} = L^{(t+1)}-L^{(t)}\approx 
    \frac{\partial L^{(t)}}{ \partial\theta^{ij}} \left(\theta_{ij}^{(t+1)}-\theta_{ij}^{(t)} \right)
    =-\eta 
    \left(
    \frac{ \partial L^{(t)}} {\partial\theta^{ij}}\right)^2. 
\end{align}
In the linear case, this simplifies to the trace of the global Hessian, multiplied by the loss, as shown in \cref{sec:timescale}. We demonstrate that a similar behavior holds for a realistic DNN in \cref{fig:probe}(center, right).

To show how this relates to the noise node, it is easier (though not necessary) to take the continuous time limit ($\eta \to 0$) of~\cref{eq:linear_evolution} resulting in a set of coupled Langevin equations~\cite{doi:10.1119/1.18725}.
In~\cref{sec:timescale}, we show that if we can neglect the effect of the NIN on the loss, the solution of the full continuous time equations leads to the following relations
\begin{align}
    L_{\epsilon=0}(t)
    &\sim
    L_{\epsilon=0}(0)
    e^{-2 \sigma_x^2 
    \int_0^t \left[ (w^{(0)}(t'))^2+ (w^{(1)}(t'))^2\right] dt' }
    , 
    \\ 
    \nn
w_{\mathrm{NI}}(t) 
    &\sim
      e^{-\sigma_\epsilon^2
      \int_0^t ( w^{(1)}(t'))^2 dt' }.
\end{align}

We define the two time scales
\begin{align}
\tau_\mathrm{noise} =-w_\mathrm{NI}/\dot{w}_\mathrm{NI}|_{t=0}
      ,
      \qquad
      ~\tau_\mathrm{loss}=-L_{\epsilon=0}/\dot{L}_{\epsilon=0}|_{t=0},
\end{align}
which differ both by a multiplicative factor and by the magnitude of the weight connecting the input with the first layer.
The first comes from the difference between data and noise distributions, while the second is due to the NINs being connected at the same layer as the input ($\ell_\mathrm{NI}=0$), such that the local Hessian which controls their evolution is insensitive to first layer contributions to the loss evolution.
Adding multiple NINs in more complicated constructions which would be sensitive to the input layer's weights could provide a novel probe for how various local parts of the Hessian affect the evolution of loss functions.

\section{Conclusions}
\label{sec:broad}

The study of the relations between training dynamics and learning capabilities of DNNs via the injection of noise through optimizable connections provides a novel path toward interpretable deep learning.
In the future, NINs could have many possible technological applications. 
For instance, one could recognize the evolution of NIWs among regular data weights to identify features which are uncorrelated with the labels, and thus build normalization schemes~\cite{Ioffe2015BatchNA, https://doi.org/10.48550/arxiv.1603.01431,Hoffer2018NormME,Schneider2020ImprovingRA,Xu2021WhySL} to suppress such weights in an effort to tackle internal covariate shift.
Furthermore, it's possible that by observing the evolution of the NIWs one could learn more about how the different parts of the DNN affect the evolution of the loss, as well as possibly teach one about generalization, as the NIWs cannot be memorized by the network, only suppressed.
Specifically, landing in different points on the loss landscape as a result of noise mitigation could inform us on the possible relation between loss curvature and noise resistant networks.
While this early study clearly shows that our analytical understanding captures at least some features of NIN dynamics, more complicated architectures are expected to have much richer structure and thus more complex evolution.
For example, bounded nonlinearities cannot generate a divergence in the loss function, implying a different behavior during the catapult and divergent phases.
There is also a clear interplay with the network hyperparameters --- a small batch size increases the effect of the stochastic gradient, while changing the learning rate can change the boundaries of the phases.
Analytical studies of other DNNs, including convolutional, recurrent, and transformer architectures, could identify analogous probes for techniques deployed at industry scale.
%
% Insights on the inner workings of DNNs naturally lead to technological improvements, raising concerns regarding potentially harmful future applications.
% %
% While training with a NIN may result in more robust networks with respect to various forms of input distortions, it encourages the interplay of such stabilities with biases present in the training sets of many real-world applications (\eg, biased curation of images or corpora for natural-language processing) to be explored in future research.
% %
% However, we believe the self-probing nature of NIWs can be used to successfully interrogate the difference between such effects, and provide an overall positive impact on network training.

\section*{Acknowledgments}
We thank Yasaman Bahri, Yohai Bar-Sinai, Kyle Cranmer, Guy Gur-Ari and Niv Haim for useful discussions and comments at various stages of this work.
NL would like to thank the Milner Foundation for the award of a Milner Fellowship.
MF is supported by the DOE under grant DE-SC0010008.
MF would like to thank Tel Aviv University, the Aspen Center for Physics (supported by the U.S. National Science Foundation grant PHY-1607611), and the Galileo Galilei Institute for their hospitality while this work was in progress.
The work of TV is supported by the Israel Science Foundation (grant No. 1862/21), by the Binational Science Foundation (grant No. 2020220) and by the European Research Council (ERC) under the EU Horizon 2020 Programme (ERC-CoG-2015 - Proposal n. 682676 LDMThExp).

\appendix

\bibliography{MLbib}

\bibliographystyle{unsrtnat}

\newpage
\section{Additional Empirical Results}
\label{apd:more_exp}

Here, we present additional results not included in the main text. 

In \cref{fig:fmnist_phases_CE} below, we show the different phases predicted in \cref{sec:theory}, for a 3 layer MLP with ReLU activations, trained on FMNIST using cross-entropy (CE) loss. These results are similar to the ones obtained for a network which was trained using the MSE loss, shown in \cref{fig:linear_phases}(bottom), demonstrating the persistence of the phase structure regardless of the loss function.
%%%%%%%% Figure 3 %%%%%%%%%%%
\begin{figure}[htbp!]
  \centering
           {
                \label{subfig:FMNIST_phase_h}
             \includegraphics[width=1\linewidth]{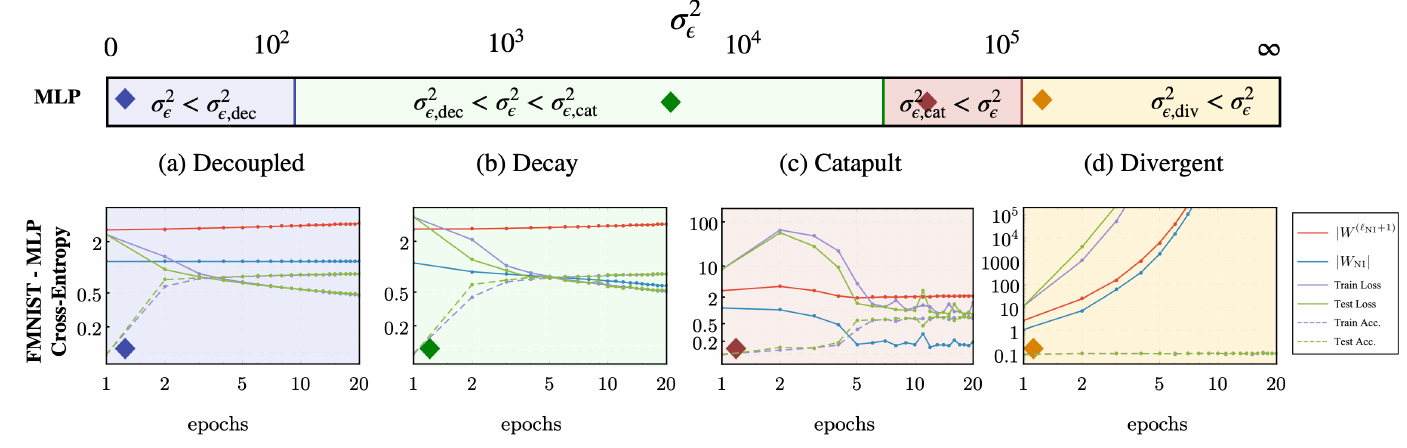} % .png .jpg ... according to supported graphics files
            }       %
  \caption{
  % FMNIST, 1 hidden layer, ReLU activations.
  Dynamical evolution of a network with a NIN coupled to the first hidden layer of a DNN. 
  {\bf Top bar:} The {\bf light shaded} regions separating the four phases as a function of the noise strength, $\sigma_\epsilon$.  The colored {\bf diamonds} indicate the values of $\sigma_\epsilon$ used in the bottom plots. 
   {\bf Bottom plots}: Various phases of the NIW dynamics during training for a 1-hidden layer MLP with Cross-Entropy loss and ReLU activations, trained on the full FMNIST dataset.
  The data weights ({\bf black} and {\bf red lines}), NIW ({\bf blue}) and loss functions ({\bf purple}) are shown.
  In the decoupled and decay phases, data weights approach optimum values while the NIW decays slowly.   
  Larger noise implies  longer time for this process to end.
  In the initial stages of the catapult phase no learning is achieved until the NIW sufficiently decays, when standard learning is resumed.  
  Lastly, in the divergent phase the system fails to learn, reaching a breakdown after only a few epochs.
{\bf Solid} ({\bf dashed}) curves represent the loss (accuracy) values, for training ({\bf violate}) and test ({\bf green}) instances.  
 The behavior displayed by the loss function, the norms of the NIWs ({\bf blue}) and the weights connected to the subsequent layer ({\bf red}), corroborates the predictions of the linear model, as well as the results obtained on the same network, trained using an MSE loss function, discussed in \cref{sec:theory}.
 For experimental details, see \cref{apd:architecture_details}.
  \label{fig:fmnist_phases_CE}
  }
\end{figure}

\section{Experimental Details}\label{apd:details}
\label{apd:architecture_details}

In the main text we present results for models trained on the Fashion-MINST (FMNIST) dataset~\citep{xiao2017/online}. FMNIST contains 70,000 grayscale images in 10 categories. The images show individual articles of clothing at low resolution ($28\times 28$ pixels). 
We preprocess the data by subtracting the mean and dividing by the variance of the training data, and train using a $60/40$ training/validation split. 
All test accuracy evaluations are done with the NIN output set to $0$, \ie,  $\epsilon = 0$.

{\bf {Implementation Details} }: For all of our experiments, we utilize a Multi-Layer Perceptron (MLP).

We optimize using vanilla SGD with either MSE or Sparse Cross-Entropy loss. The model parameters $\bs \theta, \bs w_{\rm{NI}}$ are initialized at iteration $t = 0$ using a normal distribution as $\bs{w}_0, \bs{w}_{\rm{NI},0} = 1/\sqrt{N_{\mathrm{fan-in}}}$ unless otherwise specified. 

Here we describe experimental settings specific to a figure.

\Cref{fig:probe}.
Fully connected, 3 hidden layers, $N_w = 1024$, ReLU non-linearity trained using SGD (no momentum) on FMNIST using a Sparse-Cross-Entropy loss function. Batch size = 128, with learning rate $\eta=0.05$, using weight normalization $w^{(\ell)}\sim \mathcal{N}(0,1/d_\ell), b = 0$.  The hyperparameters are chosen to obtain good generalization performance without a NIN.

\Cref{fig:linear_phases}. (top row):
Numerical solution of \cref{eq:linear_evolution}. Learning rate set at $\eta=0.01$. Intialization values for the weights were chosen to mimic a standard over-parameterized regime, i.e. $w^{(0)},w_{\rm{NI}}^{(0)}\sim \mathcal{N}(0,1)$, while $w^{(0)},w_{\rm{NI}}^{(0)}\sim \mathcal{N}(0,1/100)$. The target is the identity, i.e. $M=1$, as the equations are invariant under reparameterization with respect to $M$. The only values of consequence are then the ratios of weights at initialization and the learning rate. Therefore the actual values of the learning rate and noise variance are chosen to make the effects visible to the eye for a short training period but do not affect the final results in any way.

\Cref{fig:linear_phases} (bottom row), \cref{fig:fmnist_phases_CE}. Fully connected, one hidden layer $N_w = 1024$, ReLU non-linearity trained using SGD (no momentum) on FMNIST. Batch size = 1000, with learning rate $\eta=0.01$, using weight normalization $W^{(\ell)}\sim \mathcal{N}(0,1/d_\ell), b = 0$.  The hyperparameters are chosen to obtain good generalization performance without a NIN.

In \cref{fig:linear_phases}(bottom) and \cref{fig:fmnist_phases_CE} we use different loss functions, namely, MSE and Cross-Entropy, respectively. We detail the amount of noise injection used in each of the two figures in~\cref{tab:noise_strength}.

\begin{table}[]
\centering
\begin{tabular}{|l|l|l|l|l|}
\hline
    & Decoupled  ($\sigma_\epsilon^2$)                  
    & Decay  ($\sigma_\epsilon^2$)
    & Catapult   ($\sigma_\epsilon^2$)                      & Divergent   ($\sigma_\epsilon^2$)                        \\ \hline
MSE & $10^{-4}\cdot d_{\rm{Input}}/(N_w \eta)$ & 
$30\cdot d_{\rm{Input}}/(N_w \eta)$ & 
$47 \cdot d_{\rm{Input}}/ (N_w \eta)$ & 
$50 \cdot d_{\rm{Input}}/ (N_w \eta)$ \\ \hline
CE  & $10^{-4} \cdot d_{\rm{Input}}/ \eta $    & 
$0.1\cdot d_{\rm{Input}}/ \eta$     & 
$d_{\rm{Input}}/ \eta$                & 
$1.8\cdot d_{\rm{Input}}/ \eta      $ \\ \hline
\end{tabular}
\caption{Noise injection strength ($\sigma_\epsilon^2$) used in \cref{fig:linear_phases}(bottom), denoted as MSE, and \cref{fig:fmnist_phases_CE}, denoted as CE.
\label{tab:noise_strength}}
\end{table}

\section{Further Details On the Noise Parameter Expansion}
\label{apd:epsilon}

Here we provide additional details on the theoretical analysis of the general model in \cref{sec:theory}.
Starting with the noise translated loss function
\begin{align} \nn
    L({\bs\theta}, {W_\mathrm{NI}})
    &= 
    \frac{1}{|\mb|} \sum_{\{\bs x, \epsilon, \bs y\}\in \mb}
        \La(\bs \theta, { W_{\rm{NI}}}; {\bs{x}},\epsilon, \bs{y})
    = \frac{1}{|\mb|} \sum_{\{\bs x,\epsilon, \bs y \}\in \mb} 
        e^{\epsilon { W^T_{\rm{NI}}}\nabla_{{\bs z}^{(\ell_{\rm NI})} } }
        \La(\bs \theta; {\bs{x}},  \bs{y})
    \\ \nn
        &=
    L({\bs{\theta}})
      + \frac{1}{|\mb|} 
      \sum_{\substack{ \{\bs x, \epsilon, \bs y\}
    \\\in \mb}}
          \sum_{k=1}^{\infty}
          \frac{1}{k!}
            (\epsilon { W^T_{\rm{NI}}} \cdot \nabla_{{\bs z^{(\ell_{\rm{NI}})}}}
              )^k 
        \La(\bs \theta; {\bs{x}}, \epsilon, {\bs y}).
\end{align}
Expanding in powers of $\epsilon { W_{\rm{NI}}}$, we obtain an infinite series given by
\begin{align}
    \label{eq:NINRloss2}
  L({\bs{\theta}}, {W_{\rm{NI}}})
    = L({\bs{\theta}})
      + \frac{1}{|\mb|} \sum_{  \{\bs x, \epsilon, \bs y \} \in \mb}
          \sum_{k=1}^{\infty}
          \frac{1}{k!}
            (\epsilon { W^T_{\rm{NI}}} \cdot \nabla_{{\bs z^{(\ell_{\rm{NI}})}}}
              )^k 
        \La(\bs \theta; {\bs{x}}, \epsilon, {\bs y}).
\end{align}
Performing the batch averaging explicitly amounts to averaging over the statistics of the data and noise, resulting in the loss
\begin{align}\label{eq:average_loss}
    \langle \mathcal{L}(\bs \theta, W_{\rm{NI}}) \rangle
    = 
    \langle \mathcal{L}(\bs \theta) \rangle
    + W^T_{\rm{NI}} \cdot \langle { \epsilon } \bs  \grad_{\ell_{\rm{NI}}} \rangle
    +
    \frac{1}{2}  { W}_{\rm{NI}}^{T}\langle \epsilon^2 \mathcal{ H}_{\ell_{\rm{NI}} } \rangle {W}_{\rm{NI}}
    +\ldots
\end{align}
Here, batch averaging is denoted by $\langle .. \rangle$, 
the {\it local} gradient is 
${\bs{\grad} }_{\ell _{\rm{NI}}}
 =  \nabla_{{\bs z}^{(\ell_{\rm{NI}})}} \La({\bs \theta}; {\bs x}, {\bs y})$ and Hessian ${ {\mathcal{H}_{\ell_{\rm{NI}} }} }  
=  { \nabla_{\bs z^{(\ell_{\rm{NI}})}}\nabla^T_{\bs z^{(\ell_{\rm{NI}})}}} \La(\bs \theta; \bs x, \bs y)$ are network-dependent functions, pertaining to the NIN layer.

Taking $\epsilon$ sampled from a distribution with zero mean, the local gradient and Hessian contributions simplify as $\langle \epsilon {\bs \grad}_{\ell_{\rm{NI}}} \rangle \sim  \sqrt{\langle {\bs \grad}_{\ell_{\rm{NI}}}^2 \rangle} \Phi \sigma_\epsilon/ \sqrt{|\mb|}$, where $\langle {\bs \grad}_{\ell_{\rm{NI}}}^2\rangle $ is the vector of the batch-averaged absolute value of the gradient and $\Phi$ is a random variable with mean $0$ and variance $1$, and $\langle \epsilon^2 \mathcal{H}_{\ell_\mathrm{NI}} \rangle
\sim \sigma_\epsilon^2 \langle \mathcal{ H}_{\ell_{\rm{NI}} } \rangle$,
where we define $\sigma^2_\epsilon\equiv \langle \epsilon^2 \rangle$ as the variance of the injected noise. 

This result, as proven below, makes explicit that odd terms in the expansion are suppressed by $|\mb|^{-1/2}$. 

The last step in the derivation is performed by taking the SGD update step with respect to the NIWs, which is simply

\begin{align}
    W_{\rm{NI}}^{(t+1)}
    =
    W_{\rm{NI}}^{(t)}
    -\eta \frac{\partial L(\bs \theta^{(t)}, W_{\rm{NI}}^{(t)}) }{\partial W_{\rm{NI}}^{(t)}}.
\end{align}

Finally, utilizing \cref{eq:average_loss} we arrive at
\begin{align}
  W_{\rm{NI}}^{(t+1)} 
  = 
  W_{\rm{NI}}^{(t)}
              - \eta \frac{\sigma_\epsilon \Phi}
              {\sqrt{| \mb|}}
                  \sqrt{\langle (\bs\grad_{\ell_{\rm{NI}} }^{(t)})^2 \rangle}
              - \frac{\eta\sigma^2_\epsilon}{2}
                   \left\langle \mathcal{H}_{\ell_{\rm{NI}}}^{(t)} \right\rangle W_{\rm{NI}}^{(t)}
                   +\dots ,
\end{align}

which is given in the main text as~\cref{eq:newweight}.

We now turn to discuss our estimation of batch-averaged terms which are proportional to powers of $\epsilon$. Generically, such terms can be written as $q \epsilon^n$, where $n$ is an integer, and $q$ is some sample-dependent variable, possibly with other indices.

The goal of this appendix is to prove the following theorem:
\begin{thm}
Let q be a sample-dependent variable, with a finite mean $\langle q\rangle$ and with $Q\equiv\sqrt{\langle q^2 \rangle}$. And let $\epsilon$ be the output of some NIN, which has a PDF symmetric around zero (and consequentially of zero mean), and for some integer $n$, $\langle\epsilon^{n}\rangle=(\sigma_{n,\epsilon})^{n}$, and $\langle\epsilon^{2n}\rangle=(\sigma_{2n,\epsilon})^{2n}$. For such a case, the average of $q\epsilon^n$ over a batch ${\mathcal{B}}$ has a mean of $\langle q\rangle (\sigma_{n,\epsilon})^{n} $, and a variance of $\left(\sigma_{2n,\epsilon}^{2n} Q^2-\langle q \rangle^2\sigma_{n,\epsilon}^{2n}\right)/\sqrt{|B|}$.
\end{thm}

For odd $n$s, this simplifies to having zero mean, and a standard deviation of $Q\sigma_{2n,\epsilon}^n/\sqrt{|\mathcal{B}|}$. 

We note that this theorem is the reasoning for the estimate $|\left\langle q \epsilon^n\right\rangle_{\mathcal B}|\sim \mathcal{O}(1) (\sigma_{\epsilon,2n})^n Q/\sqrt{|\mathcal{B}|}$ for odd $n$s, as well as the estimate of $\left\langle q \epsilon^n\right\rangle_{\mathcal B}\approx \sigma_{n,\epsilon}^{2n} \left\langle q\right\rangle$, for an even $n$ and a large enough $|\mathcal{B}|$. We also note that for a large enough $|\mathcal{B}|$, this theorem immediately follows from the law of large numbers, however we prove it for any $|\mathcal{B}|$.

\begin{proof}
We can see that for a single term, and the average over the entire distribution
\begin{equation}
        \left\langle q\epsilon^n\right\rangle=\left\langle q\right\rangle\left\langle\epsilon^n\right\rangle.
\end{equation}
This is due to the fact that $\epsilon$ is a random output that is independent of $q$, and thus $\langle q \epsilon^n \rangle=\langle q \rangle\langle\epsilon^n \rangle$. For an odd $n$, this is simply equal to 0, since $\epsilon$'s PDF is symmetric around zero, so for odd $n$s, $\langle\epsilon^n \rangle=0$.

Similarly, we may note that
\begin{equation}
        {\rm Var}( q\epsilon^n)=\left\langle(q\epsilon^n)^2-\left\langle(q\epsilon^n)\right\rangle^2\right\rangle=Q^2\sigma_{\epsilon,n}^{2n}-\langle q\rangle \langle \epsilon^n\rangle.
\end{equation}
Where we once again use the independence of $q$ and $\epsilon$, and this time also use the definitions of $Q$ and $\sigma_{\epsilon,n}$. 

We can now see that
\begin{equation}
\label{eq:meananswerodds}
    \left\langle \left\langle q\epsilon^n\right\rangle_{\mathcal{B}} \right\rangle=\frac{1}{|\mathcal{B}|}\left\langle\sum_{i=1}^{|\mathcal{B}|}q_i\epsilon_i^n\right\rangle=\frac{1}{|\mathcal{B}|}\sum_{i=1}^{|\mathcal{B}|}\left\langle q\epsilon^n\right\rangle=\langle q\rangle \langle \epsilon^n\rangle,
\end{equation}
and similarly, 

\begin{equation}
\label{eq:varanswerodds}
    {\rm Var}\left(\left\langle q\epsilon^n\right\rangle_{\mathcal{B}}\right)=\left\langle \left\langle q\epsilon^n\right\rangle_{\mathcal{B}}^2-\left\langle q\epsilon^n\right\rangle_{\mathcal{B}}^2 \right\rangle.
\end{equation}
We have already computed the second term, so now let us compute the first one,

\begin{equation}\label{eq:varbatch}
\left\langle \left\langle q\epsilon^n\right\rangle_{\mathcal{B}}^2 \right\rangle=
    \frac{1}{|\mathcal{B}|^2}\left\langle\sum_{j=1}^{|\mathcal{B}|}\sum_{i=1}^{|\mathcal{B}|}(q_j\epsilon_j^n)(q_i\epsilon_i^n)\right\rangle=
    \frac{1}{|\mathcal{B}|}\left\langle q^2\epsilon^{2n}\right\rangle+\frac{1}{|\mathcal{B}|^2}\left\langle\sum_{j=1}^{|\mathcal{B}|}\sum_{i=1,i\neq j}^{|\mathcal{B}|}(q_j\epsilon_j^n)(q_i\epsilon_i^n)\right\rangle
\end{equation},
where we divided the two sums to the contribution from $i=j$ and the contribution from $i\neq j$. The second term in Eq.~\eqref{eq:varbatch} is simply $|\mathcal{B}|^2-|\mathcal{B}|$ times the same term, and is therfore equal to,

\begin{equation}
\frac{1}{|\mathcal{B}|^2}\left\langle\sum_{j=1}^{|\mathcal{B}|}\sum_{i=1,i\neq j}^{|\mathcal{B}|}(q_j\epsilon_j^n)(q_i\epsilon_i^n)\right\rangle=\frac{|\mathcal{B}|^2-|\mathcal{B}|}{|\mathcal{B}|^2}\left\langle(q_j\epsilon_j^n)(q_i\epsilon_i^n)\right\rangle_{i\neq j}.
\end{equation}

We assume there is an arbitrarily large space of samples, which we may name as $\mathcal{A}$. Let us write its size as $|\mathcal{A}|$, and assume that $|\mathcal{A}|\to\infty$ and therefore, 

\begin{align}
    \left\langle(q_j\epsilon_j^n)(q_i\epsilon_i^n)\right\rangle_{i\neq j}
    &=\frac{1}{|\mathcal{A}|}\sum_{i,j\in \mathcal{A},i\neq j}(q_j\epsilon_j^n)(q_i\epsilon_i^n)=\frac{1}{|\mathcal{A}|^2}\sum_{i,j\in \mathcal{A}}(q_j\epsilon_j^n)(q_i\epsilon_i^n)-\frac{1}{|\mathcal{A}|^2}\sum_{j\in \mathcal{A}}(q_j\epsilon_j^n)^2
    \\ \nn
    &=\left\langle q \epsilon^n\right\rangle^2-\frac{1}{|\mathcal{A}|}\left\langle( q \epsilon^n)\right\rangle= \left\langle q \epsilon^n\right\rangle^2,
\end{align}

where in the last equality we used $|\mathcal{A}|\to\infty$. We can now write the variance of the batch averaged quantity, by collecting all the different terms, and find tha

\begin{equation}
    {\rm Var}\left(\left\langle q\epsilon^n\right\rangle_{\mathcal{B}}\right)=\frac{1}{|\mathcal{B}|}\left(\left\langle q^2\epsilon^{2n}\right\rangle-\left\langle q \epsilon^n\right\rangle^2\right),
\end{equation}
as originally postulated.
\end{proof}

\section{Decay Time-scale Derivation}
\label{sec:timescale}
Here, we derive the timescale for the NIW to decay, while the loss converges to its minimum, for the linear example given in~\cref{sec:theory}.

Assuming that the system is in the decay phase, i.e. the SGD equations are given by
\begin{align}
     w^{(1)}_{t+1} &=
                      w^{(1)}_t  - \eta (w^{(1)}_t w^{(0)}_t - M) w^{(0)}_t \sigma_x^2 
                      , \\
      w_{\mathrm{NI}, t+1} &= 
      w_{\mathrm{NI},t} (1- \eta ( w^{(1)}_t)^2 \sigma_\epsilon^2).
\end{align}
Since the noise injection does not cause the system to diverge at this stage, the continuous time limit ($\eta\to0$) is expected to hold, simplifying the equations as
\begin{align}
     \dot{w}^{(1)} &=
                        -  (w^{(1)} w^{(0)} - M) w^{(0)} \sigma_x^2 
                      , \\
      \dot{w}_{\mathrm{NI}} &= 
      - ( w^{(1)})^2 \sigma_\epsilon^2 {w}_{\mathrm{NI}},
\end{align}

where it is implied that all weights are functions of time, $w=w(t)$.
Next, we can define the loss function in the absence of noise, which will be the quantity we wish to track. This function is simply
\begin{align}
    L=\frac{1}{2}\left(   (w^{(1)} w^{(0)} - M)\sigma_x  \right)^2,
\end{align}

which results in the continuous time update equation for the loss and the data weights
\begin{align}
    \dot{L}
    &=
   \sigma_x^2 \left(   (w^{(1)} w^{(0)} - M)  \right)
   \left(  \dot{w}^{(1)} w^{(0)} + \dot{w}^{(0)} w^{(1)}   \right)
   =
   \sigma_x \sqrt{2L}\left(  \dot{w}^{(1)} w^{(0)} + \dot{w}^{(0)} w^{(1)}   \right),
   \\
    \dot{w}^{(1)}
    &=
      -{\rm{sign}} (w^{(1)} w^{(0)} - M) \sqrt{2L} \sigma_x  w^{(0)},
      \quad
       \dot{w}^{(0)}
    =
      -{\rm{sign}} (w^{(1)} w^{(0)} - M) \sqrt{2L} \sigma_x  w^{(1)},
\end{align}

combining these equations we obtain for the loss function and the NIW we have
\begin{align}
    \dot{L}
    &=
    -2 \sigma_x^2  \left( (w^{(0)})^2 + (w^{(1)})^2\right) L
    ,
    \\
\dot{w}_{\mathrm{NI}} &= 
      -\sigma_\epsilon^2 ( w^{(1)})^2  {w}_{\mathrm{NI}},
\end{align}

where we identify that the loss function evolves according to the trace of the full Hessian 
$\Tr{(H_{\bs{\theta}})} = \sigma_x^2  \left( (w^{(0)})^2 + (w^{(1)})^2\right)$, while the NIW evolves according to the trace of the local Hessian $\Tr{(\mathcal{H}_z)}=( w^{(1)})^2 $.

These equations imply an exponential evolution for both the loss function and the NIW. The relevant timescales can be read by integrating the equations, hence
\begin{align}\label{eq:loss_and_nw}
    L(t)
    &\sim
    e^{-2 \sigma_x^2 
    \int_0^t \left( (w^{(0)}(t'))^2+ (w^{(1)}(t'))^2\right) dt' }
    ,
    \\
w_{\mathrm{NI}}(t) &\sim
      e^{-\sigma_\epsilon^2
      \int_0^t ( w^{(1)}(t'))^2 dt' }.
\end{align}

as presented in the main text.

\end{document}